# A New Approach In Dynamic Traveling Salesman problem: A Hybrid Of Ant Colony Optimization And Descending Gradient

Farhad Soleimanian Gharehchopogh[1], Isa Maleki[2], Seyyed Reza Khaze[3]

[1,2,3]Computer Engineering Department, Science and Research Branch, Islamic Azad University, West Azerbaijan, Iran
{bonab.farhad[1], maleki.misa[2], khaze.reza[3]}@gmail.com

## ABSTRACT

*Nowadays swarm intelligence-based algorithms are being used widely to optimize the dynamic traveling salesman problem (DTSP). In this paper, we have used mixed method of Ant Colony Optimization (AOC) and gradient descent to optimize DTSP which differs with ACO algorithm in evaporation rate and innovative data. This approach prevents premature convergence and scape from local optimum spots and also makes it possible to find better solutions for algorithm. In this paper, we're going to offer gradient descent and ACO algorithm which in comparison to some former methods it shows that algorithm has significantly improved routes optimization.*

## 1. INTRODUCTION

Nowadays method of efficiently-solving large number of problems known as NP-Hard problems is one of the most difficult combinational optimization issues. Solving such problems means to find the best solution among a large set of solutions to the problem. DTSP is one of the most famous combinational optimization problems. This problem initially introduced by Psaraftis in 1988 [1]. In recent decades many Heuristic methods have been suggested for solving DTSP's and these methods are more flexible in comparison to the classical and traditional methods. Some heuristic methods such as combinational algorithms [2], ACO algorithms [3, 4, and 5] and evolutionary computation [6, 7] can be mentioned in this field.

Swarm intelligence methods, are tools to find optimal answers or solutions. These methods take advantage of search and cooperation to look for an optimization problem's search space. So the more one algorithm controls these parameters, more near-optimal solutions finds to the problem. ACO algorithm refers to a class of meta-heuristic algorithms that have adapted from the social behaviour of ants in nature. These algorithms' aim is to find optimal solutions in reasonable time. Ants are autonomous factors in nature and when they work together, their performance is really smart [8].

The optimization-problems' aim is to find optimal or near-optimal points. In some functions in addition to public minimum there are some other local minimums.in that case it's not possible to easily find the public optimal point. In such problems, we can use combinational algorithms in search space to find optimal solution for problem. Gradient descent is an optimization algorithm to find minimum points in function [9, 10]. Gradient descent refers to optimizing processes and adjustment of input variables affect the output of the process in them. Mathematical methods use a collection of efficient and necessary circumstances that are true in optimization problem's answer. Presence or absence of limitations in the optimization is essential in these approaches. In optimization problems without limitations such as gradient methods, the aim is to find optimal or near-optimal solution to an objective function [11, 12].

DOI : 10.5121/ijmpict.2012.3201      1

Beside swarm intelligence methods that are known as smart optimization methods, evolutionary computation or meta-heuristic methods, using mathematical methods, depending on the application may be faster. Gradient descent algorithm can process the searching in complex combinational problems space to find near-optimal solutions with high performance by creating an appropriate balance in ACO algorithms.

The overall structure of our paper is organized as follows: We have organized the general structure of this paper as following: In section second introduce literature review; in section third we introduce descent gradient algorithm; in section fourth of this paper we'll introduce ACO; in section five we'll explain the proposed algorithm, in section sixth we'll discuss about valuation and results of the proposed method and finally in section seventh we will draw some conclusions from this paper.

## 2. LITERATURE REVIEW

For more accurate optimization results, we should use pre-heuristic methods (such as ACO algorithm, Genetic Algorithm, etc.) and to optimize possible solutions, we must use computational methods (e.g. gradient descending method). As DTSP has lots of local minimums, Computational methods, start from the basic hypothetical point to find first and closest near-optimal solution, and if it find the solution, defines it as optimal point; while meta-heuristic methods Randomly in high repetitions are looking for the best solution and by jump out of point to another trying to find the best and most efficient point and provide optimal solutions. In recent years meta-heuristic methods have been considered to find near-optimal solutions to combinational optimization problems.

F.S.Gharehchopogh et al. in [2] has used ACO algorithm and genetic algorithm combinational algorithms to solve DTSP. They have used genetic algorithm in his paper to optimize paths obtained from ACO algorithm in search space. In this paper chromosome genes in genetic algorithm are paths obtained from ACO algorithm. M.Guntsch in [3] has solved DTSP using ACO algorithm. He has used probability of motion and update rules changes to optimization. As gradient descent is local and only has high potential to find minimum and maximum points, it offers probability of better and more acceptable answers to find optimal solutions to the DTSP in comparison to the meta-heuristic methods.

The increasing usage of mathematical optimization techniques has led many researchers to focus on developing and optimize problems using mathematical operations. Since there is no unique approach to solve all the optimization problems, various computational methods for solving optimization problems have emerged. Gradient descent algorithm is the basis of many optimization algorithms and it's being used to learn artificial neural networks and minimize the network errors to adjust the network parameters [13]. In [14] X.Wang has utilized gradient descent usages. He has used gradient descent in order to solve complex integrals. D.Kumar & Y.Kumar in [15] have used optimization methods. In their journal they have they have studied the optimization methods that are gained from the nature. Among these methods and ways they have marked gradient descent as an optimization method to find the max or min of the problems.

## 3. GRADIENT DESCENT ALGORITHM

There is a group of gradient algorithms in which the process of searching for gradient is used with some changes, to find the optimal answer. The continuous answers which are found in continuous repeats are similar to the main optimal answer. So, we stop when the experimental answers are gotten similar to the optimal answer much enough [16, 17].

Since there is no limitation in the problem it seems that moving in the same direction as gradient could be a practical way [17]. If the purpose of the problem is to minimize the f(x) we have to move in the opposite direction as the gradient. In the other words, we should use equation (1) in order to achieve the answer [16].

$$x_{n+1} = x_n - t_n \nabla f(x_n) \tag{1}$$

In the above equation (1) the function has been begun from an arbitrary point to find the optimal answer and the function also performs minimizing in the same direction as gradient. In the equation (1), t is a non-negative scalar which minimizes the function in the same direction as gradient. The new gradient is created by the previous one in each repeat.

When x is decreased we move from the current experimental answer to the next answer. Thus, in order to generalize a searching method for a variable, firstly we should specify the movement direction of the gradient using the partial derivatives. As we assume that f(x) is derivable, in each point x will have a gradient vector which is represented in the equation (2) as ∇f(x') [16].

$$\nabla f(x') = (\frac{\partial f}{\partial x_1}, \frac{\partial f}{\partial x_2}, ..., \frac{\partial f}{\partial x_n}) \qquad x = x' \tag{2}$$

The reason that gradient is important is that we can find minimal changes in function in x point as we move in the same direction as gradient.

Pseudo code for the algorithm of the gradient descent is represented in the following:

> Choose randomly $x_0$
> **While** ‖f($x_{n+1}$) – f($x_n$) > □‖ **do**
> Choose a decreasing $\gamma_n$ (generally 1/n)
> Compute $x_{n+1} = x_n - t_n \nabla f(x_n)$
> **End while**
> Do some random restarts
> **Return** the lowest couple $x_n$, f($x_n$) found.

Figure 1. Pseudo Code for the Algorithm of the Gradient Descent

As the purpose of the problem is to find the optimal answer, if we move in the same direction as gradient, we will achieve the answer much sooner [19]. The method of gradient is to move towards the optimized answer in a zigzag direction instead of a straight direction.

An appropriate reaction is to start the movement from the current answer in a distinct direction and not to stop as the f(x) is decreasing. When f(x) stops decreasing we stop too and the next answer is ready there. So, the gradient is calculated again to specify the direction of the next movement [19]. In this reaction each repeat contains the changing of the current answer of the X. The repeats of the gradient will continue until the gradient equals zero, if the distance of ε is considered acceptable. According to equation (3) we stopped [16].

$$\left[\frac{\partial f}{\partial x_n}\right] \leq \varepsilon \qquad n = 1, 2, ..., n \tag{3}$$

ε is the distance of the gradient. It can be assumed either short or long. If the distance is short, the convergence rate is low. If this distance is small, the convergence rate is low. Gradient with increasing distance until convergence is guaranteed, the convergence rate increases.

## 4. ANT COLONY OPTIMIZATION (ACO)

ACO algorithm, suggested by Dorigo in 1996 [20, 21], is a branch of artificial intelligence which is known as collective intelligence group. The movement probability of the ant K from the city I to the city j in the time t is suggested based on the equation (4). In this statement $\eta_{i,j}$ is the purview and equals $1/d_{i,j}$ (the more nearby cities are more likely to be chosen and $\tau_{i,j}$ is the amount of pheromone on the mane in the time t.

*Allowed k* is the group of cities that the ant hasn't met up until now but can possibly meet them at the next step. α & β are the parameters of the impact of the pheromone laid on the main and the distance impact, respectively.

$$p_{ij}^k(t) = \begin{cases} \dfrac{[\tau_{ij}(t)]^\alpha \cdot [\eta_{ij}]^\beta}{\sum_{j \in allowed_k}[\tau_{ij}(t)]^\alpha \cdot [\eta_{ij}]^\beta} & \text{if } k \in allowed_k \\ 0 & otherwise \end{cases} \quad (4)$$

The update rule of the laid pheromone on the mane is formed based on the equation (5).

$$\tau_{ij}(t+n) = (1-\rho) \times \tau_{ij}(t) + \Delta \tau_{ij} \quad (5)$$

In the equation (5), $1-\rho$ specifies the evaporation rate of the pheromone between the time t and (t + n). $\rho$ Is assumed as the parameter of the pheromone evaporation in the range that: $(0 < \rho < 1)$. The much the amount of the $\rho$ is, the much the speed of evaporation increases.

$$\Delta \tau_{ij} = \sum_{k=1}^{m} \Delta \tau_{ij}^k \quad (6)$$

$\Delta \tau_{ij}^k$ Represents the amount of pheromone of the best tour and it is calculated according to the equation (7).

$$\Delta \tau_{ij}^k = \begin{cases} Q/L_k & \text{if k ant uses edge (i, j) at time } (t, t+n) \\ 0 & otherwise \end{cases} \quad (7)$$

In the equation (7), $Q$ is a constant parameter which is dependent on the problem and $L_k$ is the length of the path which is travelled by the ant K. Thus, the initial level of pheromone and evaporation rate, an important influence on the optimization process and is required to be determined in an appropriate way.

## 5. PROPOSAL ALGORITHM

A large number of optimization methods are called collective optimization methods. In these methods algorithm starts finding solutions with some numbers of population. Then the algorithm is repeated among variables until there appears a proper answer. This process will go on until there appears an optimized answer for the problem. There are a lot's of ways in order to optimize the objective function such as numerical methods or random methods. Normally, the computational methods jump over the objective function and move toward the lower slope. Combinatorial problems are often longer and cannot be solved using computational v. the answer that is achieved on thee algorithms may be very far from the optimal answer.

In this proposed method, we use a combined method (which includes both ACO algorithm and gradient descent method) in order to optimize The DTSP. In this way, we have the high speed of gradient descent methods and also are able to search for the overall minima. The ACO algorithm like other collective intelligence algorithms has two major disadvantages. The first disadvantage is the premature convergence. Other disadvantages of the ACO algorithm are the impact of parameters of this algorithm on the problems. If the parameters are not adjusted properly, the ACO algorithm hades towards local optimization and the algorithm performance decreases. In order to avoid disadvantages of premature convergence and the dependence of algorithm parameters on the problems, we have applied some changes in the ACO algorithm. By these changes, we try to solve DTSP's by increasing the variety of solutions on how to update the pheromone. Math is used as a tool for finding optimization of many algorithms. Grad search method is one of the mathematical optimization techniques. The main idea of this method is that the optimization problem (here is finding the optimal path) begins by choosing any primary answer. Then all adjacent neighbours with the primary point are evaluated by objective function. Then between answers to neighbours and the primary point, the answer is the most acceptable is replaced to selected primary point and the algorithm is repeated until a point is found that no more than your neighbor is not acceptable algorithm stops. Fig. 2 shows the schema of the proposed algorithm.

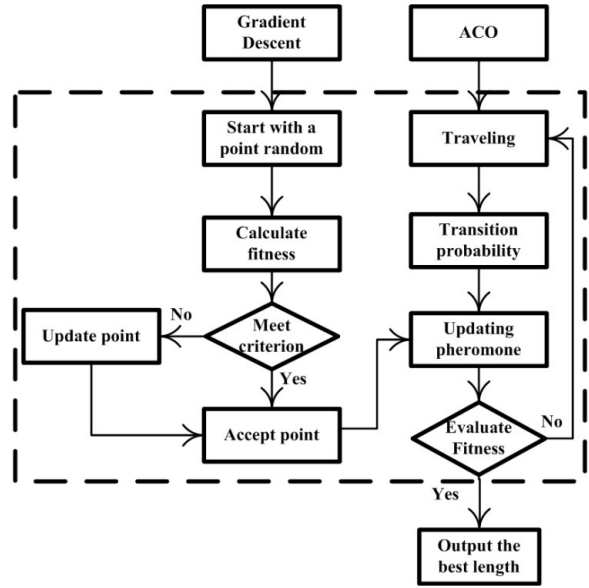

Figure 2. Flowchart of the Proposed Algorithm.

Thus, by taking the descending gradient movement, ACO have been modified in order to reach the optimal solution can be achieved more quickly. For one problem, there are several possible answers that to compare and choose the best optimal solution, a function called the objective function is defined. Select this function is dependent on the nature of the problem. In the DTSP, path length is a common goal of optimization. Pseudo code of proposed algorithm based on optimization algorithm of ACO algorithm and descending gradient includes the Fig. 3:

```
1. Initialize parameters
2. Loop
     Each ant is positioned on a starting city
  Repeat
     State transition rule
     Local pheromone updating
3. Gradient descent
     Start with a point random
    Repeat
      Determine a descent direction
      Choose a step
      Update point
    Until stopping criterion is satisfied
  Until all ants have built a complete solution
  Global pheromone updating
  Until End condition is reached.
```

Figure 3. Pseudo Code of Proposed Algorithm

To obtain the more optimal routes, descending grad algorithm is combined with the pheromone's updating law. Pheromone's updating law when a more optimal answer achieved for algorithm compared to previous repetitions, increased in optimized paths of ant pheromone.

When a better answer than previous repetitions is obtained it is possible that in the neighbourhood of this answer, exist another best answer that can be find out the answer with further searches in the neighbourhood of current answer.

As shown in Fig. 4, the algorithm works based on removing edge of the tour and reconnect it in another way. Assuming the city, is the third city, ants meet all paths connected to the third city according to descending gradient. Whatever length of the path is less, the steep of grad is more in that path, and the path is chosen. It should be noted that there are several ways to make the tour again but only the condition will be accepted that in the constraints of problem (see each city only once) applies and new tour obtained better amount than previous answer. Remove of edge and reconnect it in succession will continue as far as to not found any new improver movement for algorithm.

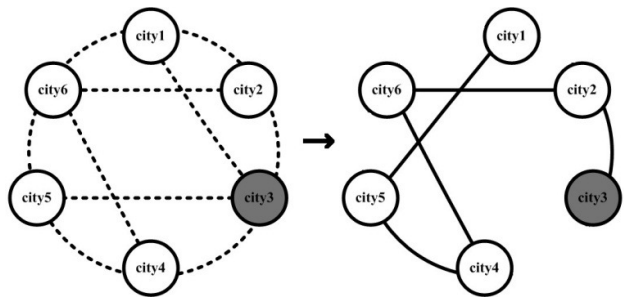

Figure 4. How to Select Cities

According to what can be seen from figure this six samples city have been obtained as a primary tour. This output of the desirable or undesirable to reduce the path length using the gradient is descending for the next tour.

To increase the efficiency of the suggested algorithm mode of pheromone updating can be combined with equation (8).

$$x_{n+1} = x_n - t_n \nabla f(x_n) \qquad (8)$$

By putting the amount of descending gradient in the right side of the equation (3) the following equation is obtained.

$$\tau_{ij}(t+n) = (1-\rho) \times \tau_{ij}(t) + \Delta \tau_{ij} + x_n \qquad (10)$$

In the above equation for select the best paths, the upper limit of pheromone is obtained.

A meta-innovative algorithm cannot reach the optimum quality of the final answer, if the parameters are not set correctly. In order to solve this problem, to compute the overall optimum, we should distinguish the problem-dependent parameters to algorithms is converge faster to the optimal answer. In this regard, we use descending grad for set pheromone In ACO algorithm. When we are faced with a large number of independent variables, determination the appropriate combination of these parameters to presence in problem and in order to estimate problem dependent variable, is importance.

Descending gradient is a firm computational tool for selecting best combination variables of problem, that can done by ascending and descending selection approach. In descending method, the amount that has the greatest impact, is added to the model. In fact in descending gradient algorithm the value that has the greatest impact is added to subject and value is less effective than the model is removed.

## 6. EVALUATION AND RESULTS

To evaluate the proposed algorithm with other algorithms, the number of cities considered to be 30 and results is estimated for 100 times iterations of the algorithm. One of the important parameters in the design of hybrid algorithm optimization and the resulting is path length. Descending gradient algorithm performance is dependent on the initial point and it does not have the ability to search all areas of problem. Thus by using ACO, that searches large space, descending grad algorithm is capable to approach the original optimal point. Parameter amounts for the proposed algorithm results are shown in Table 1.

In the proposed algorithm, there are several parameters that affect the performance, in the Table 1, by using the parameter (α) the amount of spilled pheromone on edge applied. Parameter β determine the importance of relative distance to pheromone, in the selection of the next city. Parameter (rho) weakened (evaporation) of pheromone, and the parameter (t) minimizing the function in direction of descending gradient.

Table 1. Value of Parameters in the Proposed Algorithm

| Parameter Name | α | β | rho | t |
|---|---|---|---|---|
| Value | 1 | 5 | 0.1 | 0.4 |

Experimental results show that the use of descending gradient to select path, makes a significant improvement in the efficiency and the convergence rate of the algorithm and cause the close answers to optimal.

In the proposed algorithm for increase variety in the answer, after ants converging to a path, we second initialized that. This means that the value of each ant is equal to reduction of path length that ant has wended in the previous repetition and the best length achieved by other ants.

Table 2, shows comparing proposed method with other algorithms. As the results show the proposed algorithm is able to achieve a better response than the other algorithms.

Table 2. Results obtained of the proposed algorithm and other algorithm

| Algorithms | Average Solution | Best Solution | Worst Solution |
|---|---|---|---|
| ACO [2] | 385 | 340 | 368 |
| GA [2] | 464 | 349 | 826 |
| Hybrid Algorithm [2] | 384 | 340 | 358 |
| Proposed Algorithm | 365 | 340 | 347 |

Diagram Fig. 5, comparing the proposed algorithm with other algorithms shows. As Fig. 5 shows the performance of proposed algorithm is better.

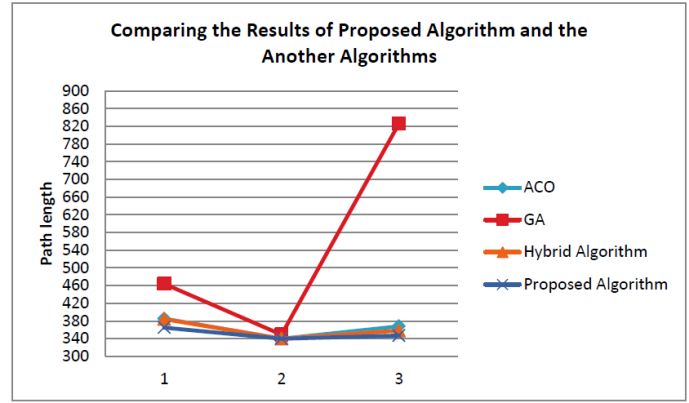

Figure 5. compared performance of the proposed algorithm with other algorithms

In Fig. 6, the impact of parameter (t) with 10 times performances in the proposed algorithm process is shown. According to Fig. 6, it is better for be optimized paths, reduce the amount of (t). With increase the amount of (t), the convergence rate has slowed down.

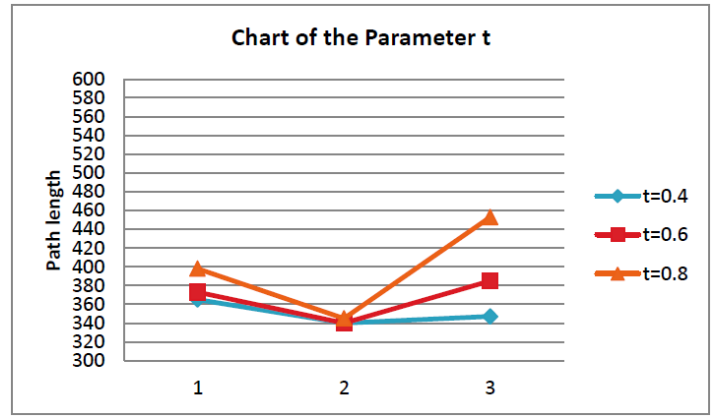

Figure 6. The impact of parameter (t)

Increasing the amount of (t) reduce the steep of descending grad. Because the possibility of production new parts increases and searching becomes more extensive. So, reducing the amount of (t) intensified the optimization process.

## 7. CONCLUSIONS

In this paper, we have described a new technique for optimizing the DTSP. In problem solving of DTSP meta-innovative methods are used mainly. In this paper a combination method from analytical techniques, using descending grad and collective intelligence methods for this problem has been proposed. Algorithm Compound of ACO algorithm and descending grad algorithm causes when caught in a local minimum, with initialized pheromone as descending grad to algorithm we cause to escape from a local minimum. The results also show that answer of the proposed method is better than the quite meta-innovative methods.

## REFERENCES

[1] H. N. Psaraftis,"*Dynamic vehicle routing problems*". In: Golden, B.L., Assad, A.A. (eds.) Vehicle Routing: Methods and Studies, Elsevier, Amsterdam, pp. 223–248, 1988.

[2] F.S. Gharehchopogh, I.Maleki, And M.Farahmandian, "*New Approach for Solving Dynamic Traveling Salesman Problem with Hybrid Genetic Algorithms and Ant Colony Optimization*", International Journal of Computer Applications (IJCA), vol:53, No.1, pp. 39-44, September 2012.

[3] M.Guntsch and M.Middendorf. "*Pheromone Modification Strategies for Ant Algorithms Applied to Dynamic TSP*". In Applications of Evolutionary Computing, Lecture Notes in Computer Science, Vol.2037, pp. 213-220, 2001.

[4] M. Guntsch, M. Middendorf and H. Schemck. "*An Ant Colony Optimization Approach to Dynamic TSP*". Proceedings of the GECCO 2001. San Francisco, USA, 7-11 July, 2001, Morgan Kaufmann, pp.860-867, 2001.

[5] C. J. Eyckelhof, M. Snoek, "*Ant systems for a dynamic TSP*". In: Proceedings of the 3rd international workshop on ant algorithms, pp. 88–99, 2002.

[6] Z.C.Huang, X.L.Hu, S.D.Chen, "*Dynamic Traveling Salesman Problem based on Evolutionary Computation*", In Congress on Evolutionary Computation, IEEE Press, pp. 1283-1288, 2001.

[7] L. Kang, A. Zhou, B. McKay, Y. Li, Z. Kang, "*Benchmarking Algorithms for Dynamic Travelling Salesman Problems*". In: Proceedings of the Congress on Evolutionary Computation, Portland, Oregon ,2004.

[8] M. Dorigo, V. Maniezzo, and A. Colorni, "*The ant system: Optimization by a colony of cooperating agents*", IEEE Transactions on System, Man, and Cybernetics, Part B, Vol. 26, pp. 29-41, 1996.

[9] G. V. Deklaits, A. Ravindran, K.M. Rogesdell, "*Engineering optimization Methods and Application*", a Wiley-Inter science Publication, 1983.

[10] A.V.Fiacco, G.P.McCormick, "*Nonlinear programming: Sequential Unconstrained Minimization Techniques*", SIAM, Philadelphia, 1990.

[11] G.P.McCormick, "*Nonlinear Programming- Theory, Algorithms and Applications*", Wiley, New York, 1983.

[12] E. Polak: "*Computational Methods in Optimization*", Academic Press, New York, 1971.

[13] R.Battiti, "*First and second order methods for learning: Between steepest descent and Newton's method*," Neural Computation, Vol. 4, No. 2, pp. 141-166, 1992.

[14] X. Wang. "*Method of Steepest Descent and its Applications*", In IEEE Microwave and Wireless Components Letters, ISSN: 15311309, Vol. 12, pp. 24-26, 2008.

[15] Y. Kumar, D. Kumar, "*Parametric Analysis of Nature Inspired Optimization Techniques*", International Journal of Computer Applications (IJCA), Vol. 32, No.3, pp. 42-49, October 2011.

[16] F. S. Hillier, G. J. Lieberman: "*Introduction to operations research*", McGraw-Hill, 2001.

[17] A. P. Engelbrecht: "*Computational intelligence*", 2nd ed, Wiley, 2007.

[18] *H. B. Curry*, *"The method of steepest descent for non-linear minimization problems"*. Quart. Appl. Math., vol.2 ,pp.258-261.1944.

[19] E.K.P. Chong, S.H. Zak: "*An Introduction to Optimization",* Wiley, New York, 1996.

[20] M. Dorigo, V. Maniezzo, and A. Colorni, *"The ant system: Optimization by a colony of cooperating agents*", IEEE Transactions on System, Man, and Cybernetics, Part B, Vol. 26, pp. 29-41, 1996.

[21] M. Dorigo and L. M. Gambardella, *"The colony system: A cooperative learning approach to the traveling salesman problem"*, IEEE Transactions on Evolutionary Computation, Vol. 1, No.1, April, 1997.